\documentclass[10pt,twocolumn,letterpaper]{article}

\usepackage{cvm}
\usepackage{times}
\usepackage{epsfig}
\usepackage{graphicx}
\usepackage{amsmath}
\usepackage{amssymb}
\usepackage{multirow}
\usepackage{bbding}
\usepackage{overpic}
\usepackage{authblk}

\usepackage[pagebackref=true,breaklinks=true,letterpaper=true,colorlinks,bookmarks=false]{hyperref}

\cvmfinalcopy

\def\cvmPaperID{****} 

\ifcvmfinal\fi
\begin{document}

\title{High-Fidelity Point Cloud Completion \\
	with Low-Resolution Recovery and Noise-Aware Upsampling}

\author[1,2]{Ren-Wu Li}
\author[3]{Bo Wang}
\author[1,2]{Chun-Peng Li}
\author[1,2]{Ling-Xiao Zhang}
\author[1,2]{Lin Gao}
\affil[1]{Institute of Computing Technology, Chinese Academy of Sciences}
\affil[2]{University of Chinese Academy of Sciences}
\affil[3]{Tencent America}

\maketitle

\begin{abstract}

Completing an unordered partial point cloud is a challenging task. Existing approaches that rely on decoding a latent feature to recover the complete shape, often lead to the completed point cloud being over-smoothing, losing details, and noisy. Instead of decoding a whole shape, we propose to decode and refine a low-resolution (low-res) point cloud first, and then performs a patch-wise noise-aware upsampling rather than interpolating the whole sparse point cloud at once, which tends to lose details. Regarding the possibility of lacking details of the initially decoded low-res point cloud, we propose an iterative refinement to recover the geometric details and a symmetrization process to preserve the trustworthy information from the input partial point cloud. 
After obtaining a sparse and complete point cloud, we propose a patch-wise upsampling strategy. 
Patch-based upsampling allows to better recover fine details unlike decoding a whole shape, however, the existing upsampling methods are not applicable to completion task due to the data discrepancy (i.e., input sparse data here is not from ground-truth). 
Therefore, we propose a patch extraction approach to generate training patch pairs between the sparse and ground-truth point clouds, and an outlier removal step to suppress the noisy points from the sparse point cloud.
Together with the low-res recovery, our whole method is able to achieve high-fidelity point cloud completion. 
Comprehensive evaluations are provided to demonstrate the effectiveness of the proposed method and its individual components. 

\end{abstract}
\section{Introduction}
\label{sec:intro}

Analyzing and understanding the 3D world \cite{dai2017shape,qi2017pointnet,meng2019vv,chen2018deep,thomas2019kpconv,li2018so,qi2017pointnet++,tchapmi2019topnet,dgcnn,hassani2019unsupervised,zhao20193d,li2018pointcnn,jin20203d} plays an increasingly important role in the rapid development of autonomous driving, robotics, and 3D reconstruction \cite{geiger2012we,rusu20113d,fan2017point}. Among different 3D data representations, point cloud data has received much attention.
However, raw point clouds are usually incomplete, sparse, and noisy due to occlusion and limitations of the acquisition sensors. Therefore, 3D point cloud completion that fixes these issues becomes a critical step for point cloud data to be successfully used in real-world applications.

Existing generative models \cite{achlioptas2017latent_pc,yang2018foldingnet,lim2019convolutional,pointflow,ShapeGF} for point cloud show promising results in learning the distribution of data and synthesizing general shapes,  but these methods are not suitable for completing point cloud. Earlier point cloud completion works \cite{shen2012structure,sung2015data} adopt search and assembly strategy, which are limited to be used for specific categories of object. Deep learning based methods have been developed lately and show promising potential for shape completion task. One group of the methods \cite{yuan2018pcn,tchapmi2019topnet,liu2020morphing,Wang_2020_CVPR} proposes to learn a global shape representation, therefore leads to the generated point cloud being over-smooth, missing fine-grained details. A second group \cite{sarmad2019rl,huang2020pf,wen2020point} aims to generate complete point cloud at low resolution due to large memory consumption limitation. Similarly, voxel or implicit field representations based shape completion methods \cite{dai2017shape,xie2020grnet,liu2019high} have resolution and memory limitations as well due to the involved 3D convolution.

In this paper, we aim to recover \textit{complete} and \textit{dense} point cloud from raw unordered partial point cloud data. \textit{Complete} means the recovery of details, and \textit{dense} indicates a high density of the produced points. Instead of decoding a complete and dense point cloud from a latent feature \cite{yuan2018pcn,tchapmi2019topnet}, which is a difficult problem to solve, as a result, losing details and increasing noise, we propose to decode a complete and sparse point cloud, refine and symmetrize it, and then perform a patch-wise noise-aware upsampling to get a final complete and dense point cloud, where each step in our method solves an easier problem. In this way, our method can achieve a better dense point cloud completion via solving several easier tasks rather than a difficult one. Our motivation is to break down the hard dense point cloud completion problem into several more tractable subproblems.

The first stage of our framework is to recover a low-res point cloud. 
In this stage, we first use an auto-encoder network to complete input partial point cloud. Due to the difficulty of recovering a complete shape from a partial one, the obtained completed sparse point cloud inevitably loses geometric details. To better recover the details of the low-res point cloud, we propose to perform an iterative refinement via residual learning, followed by a symmetrization process to preserve the trustworthy information from the input partial point cloud. After our proposed refinement and symmetrization, the updated low-res point cloud demonstrates clearer shape, more details, and less noise, which serves as input for the next upsampling stage.

The second stage of the proposed framework is to upsample the low-res point cloud obtained from the first stage. Although existing upsampling methods demonstrate promising result in increasing the resolution of point cloud from low to high \cite{yifan2019patch,li2019pu,li2021dispu}, they require input low-res point cloud to be clean and noise-free. This requirement prevents the existing methods to be applied in point cloud completion, as in point cloud completion even after careful treatment, the obtained low-res point cloud are not comparable against its corresponding ground-truth (GT) counterpart in terms of details and noise. Thus, we propose a patch-wise noise-aware upsampling method designed to be able to take imperfect and noisy low-res point cloud patch as input and produce high-fidelity high-res point cloud patch as output. Our proposed upsampling method is capable of ensuring the patches selected between the low-res and GT point clouds overlapping with each other and robustly handling noisy and imperfect low-res point cloud by filtering out the noisy points. Comparisons are provided to show the high-res point cloud obtained with the proposed strategies is significantly better than the state-of-the-art upsampling method \cite{li2021dispu}. 

In summary, our contributions are listed as follows. 
\begin{itemize}
    \item We propose to break down the difficult point cloud completion problem into low-res recovery, refinement, and patch-wise upsampling, where we solve each of them to achieve high-fidelity point cloud completion.
    \item In low-res recovery stage, we propose an iterative refinement to enchance the details and suppress noise, and a symmetrization process to preserve the trustworthy information from the input partial point cloud.
	\item In upsampling stage, we propose patch-wise noise-aware upsampling to ensure the overlap between low-res and ground-truth point cloud patches, which robustly handles the noisy and imperfect low-res point cloud.
\end{itemize}

\begin{figure*}[t]
    \centering
    \begin{overpic}[width=\textwidth]{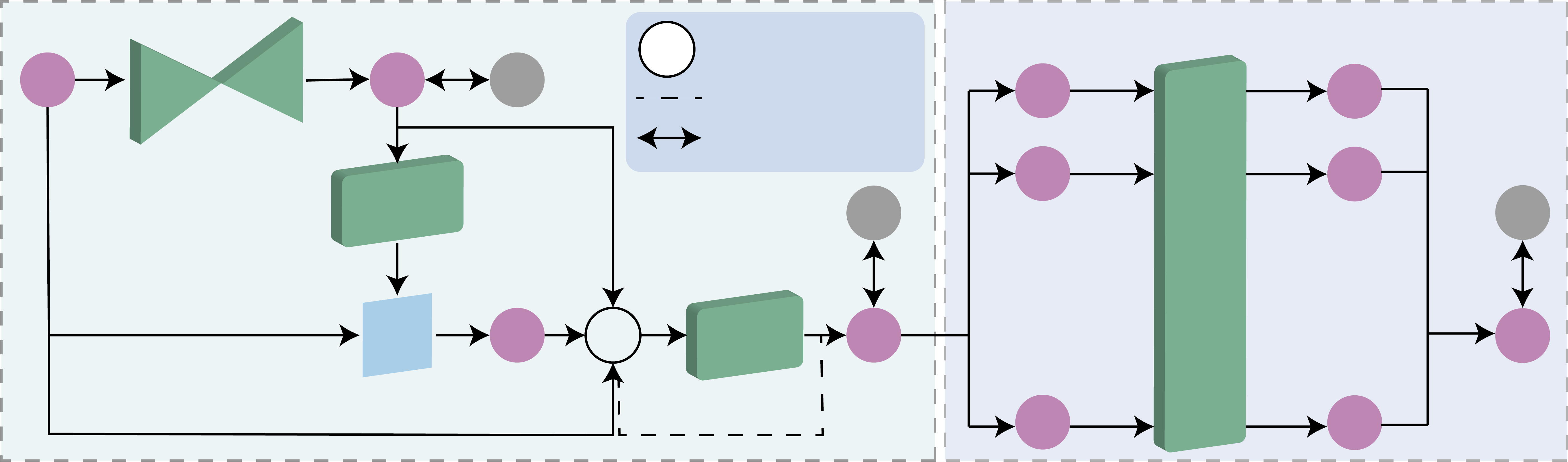}
     \put (2.25,23.8) {$\mathcal{P}$}
     \put (32.1,23.8) {$\hat{\mathcal{S}}$}
     \put (24.6,23.8) {$\mathcal{S}$}
     \put (22.8,16) {\small SymNet}
     \put (24.1,7.45) {$\mathbf{SP}$}
     \put (32,7.45) {$\mathcal{P}'$}
     \put (38.2,7.45) {$\mathrm{C}$}
     \put (45,7.45) {\small ResNet}
     \put (54.9,7.45) {$\tilde{\mathcal{S}}$}
     \put (54.9,15.2) {$\hat{\mathcal{S}}$}
     \put (65.2,23.3) {$\mathcal{A}_1$}
     \put (65.2,18.1) {$\mathcal{A}_2$}
     \put (65.2,2.2) {$\mathcal{A}_k$}
     \put (75.5,17.5) {\rotatebox{-90}{Up-Sampler}}
     \put (84.8,23.3) {$\mathcal{A}_1^{\mu}$}
     \put (84.8,18.1) {$\mathcal{A}_2^{\mu}$}
     \put (84.8,2.2) {$\mathcal{A}_k^{\mu}$}
     \put (96.5,15.2) {$\hat{\mathcal{C}}$}
     \put (96.5,7.45) {$\Tilde{\mathcal{C}}$}
     \put (41.75,25.75) {$\mathrm{C}$}
     \put (45.9,25.75) {Concatenate}  
	 \put (45.9,22.75) {Iterative}
	 \put (45.9,20) {Geometry Loss}
	\end{overpic}
    \caption{Overview of our framework. We propose to do a low-res point cloud recovery first, and follow by a patch-wise noise-aware upsampling. Given a sparse partial point cloud $\mathcal{P}$, a low-res complete point cloud $\mathcal{S}$ is firstly recovered. With the help of our symmetry detection module, SymNet, a plane of symmetry $\mathbf{SP}$ is predicted to add the symmetrical points $\mathcal{P}^{'}$ from input to $\mathcal{S}$. Afterwards, the geometry is further refined as $\Tilde{\mathcal{S}}$ by our iterative residual refinement network, ResNet. Subsequently, $\Tilde{\mathcal{S}}$ is split into overlapped patches $\mathcal{A}$ in a ball query fashion. Patch-based upsampling is then applied to each patch $\mathcal{A}_i$ to generate the final dense completed point cloud $\Tilde{\mathcal{C}}$.}
    \label{fig:pipeline}
\end{figure*}
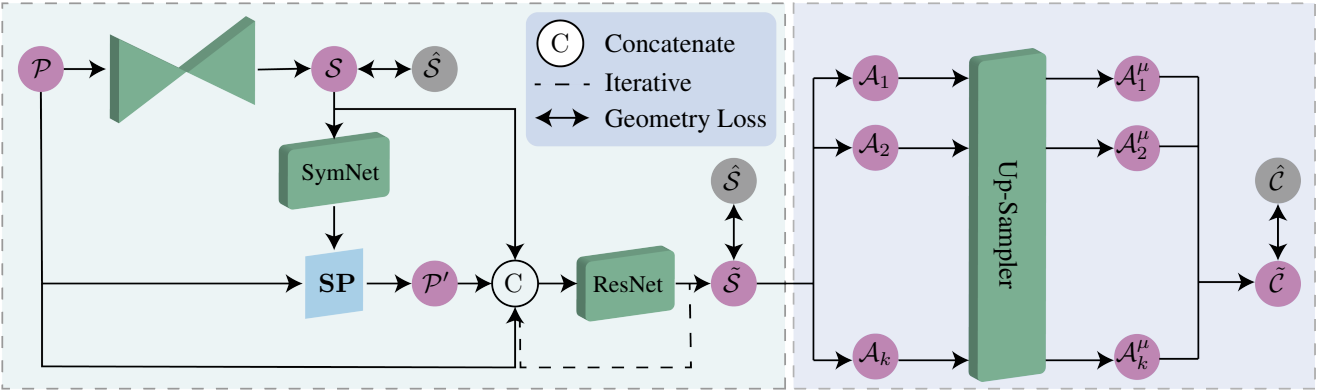

\section{Related Works}
\label{sec:related_work}

\paragraph{Point Cloud Generation.}
Generative models for point cloud have been intensively studied in recent years. 
Fan et al.~\cite{fan2017point} propose the first point set generation framework from an input image. 
Early works have proposed generative models by leveraging generative adversarial network (GAN)~\cite{goodfellow2014gan} or variational autoencoder (VAE)~\cite{kingma13vae} on point cloud generation~\cite{valsesia2018learning,achlioptas2017latent_pc}.
Achlioptas et al.~\cite{achlioptas2017latent_pc} propose to train a GAN in the latent space of a pre-trained auto-encoder.
FoldingNet~\cite{yang2018foldingnet} introduce a folding-based decoder that deforms a 2D grid to the target point cloud.
Built upon~\cite{yang2018foldingnet}, Lim et al.~\cite{lim2019convolutional} leverage the advances in image synthesis and use adaptive instance normalization to predict density and point existence locally.
PointFlow~\cite{pointflow} and ShapeGF~\cite{ShapeGF} introduce normalization flow and gradient of density field respectively to learn the distributions of shapes and points on shapes. 
Despite the progress in 3D generative models, these methods are limited to synthesizing general shapes but not applicable for shape completion task.

\vspace{-1em}
\paragraph{Shape Completion.}
There have been quite a few works focusing on shape completion since it is a fundamental and important task. Earlier works~\cite{shen2012structure,sung2015data} propose to search and assembly parts or estimate part distribution and symmetries of 3D shapes from database prior for shape completion. Recently, deep learning based point cloud completion methods have been proposed. PCN~\cite{yuan2018pcn} and MSN~\cite{liu2020morphing} learn a global shape representation by shared multilayer perceptron (MLP) to directly regress 3D coordinates of reconstructed points in an end-to-end manner, which usually lead to the generated point cloud being over-smoothing and missing details. TopNet~\cite{tchapmi2019topnet} adopts a tree-structured network to facilitate completing from partial input point cloud. Want et al.~\cite{Wang_2020_CVPR} propose a cascading refinement strategy to recover complete point cloud. 
However, as the completion task is challenging to regular autoencoder or MLP where these methods are based on, the resulting global shape representation of these methods tends to generate over smooth completion result that lacks of detailed geometry.

RL-GAN-Net~\cite{sarmad2019rl} introduces reinforcement learning agent to better enhance the ability of GAN generator to produce more realistic point cloud. PF-Net~\cite{huang2020pf} uses a self-supervised strategy and multi-stage hierarchical network to randomly drop part of a complete point cloud and predict the complete one. Wen et al.~\cite{wen2020point} adopt a similar strategy where hierarchical encoder-decoder is used. However, these approaches can only produce sparse point cloud due to its large memory cost and low efficiency.  
Voxel or implicit field representations are proposed for shape completion as well~\cite{xie2020grnet,han2017high,liu2019high,dai2017shape,stutz2020learning,chen2019learning}. However, they have the same memory/resolution issue due to 3D convolution operation, and these methods are complicated and less flexible to use.

Recent works explore alternative approaches for point cloud completion. SoftPoolNet~\cite{softpool} proposes soft pooling and regional convolution operator to improve the encoder-decoder architecture for point cloud completion and classification tasks. GRNet~\cite{xie2020grnet} introduces 3D grids as intermediate representation to regularize unordered point cloud and propose differentiable layers to convert point cloud to 3D grids. PMP-Net~\cite{Wen_2021_CVPR} points out the generation of a whole complete point cloud leading to degradation, and proposes to solve completion from a point cloud deformation perspective. NSFA~\cite{zhang2020detail} also notes decoding a whole complete point cloud is difficult, and propose to split the whole point cloud into missing and known parts and aggregate different features for each part. This is similar but inferior to our approach because they only split the whole shape into two parts, where we perform patch-wise upsampling.

\paragraph{Point Cloud Upsampling.}
Apart from doing shape completion, upsampling methods have been developed for point cloud processing~\cite{yu2018ec, yu2018pu, li2019pu}, which is related to point cloud completion since we want to produce a dense point cloud with completed shape. However, these methods are not applicable for completion task due to its requirement such that input point cloud must be complete and sparse (i.e., from ground truth not recovered) and upsampling ratio must be small. Recently by leveraging a coupled representation of voxel and implicit function, IF-Net~\cite{chibane20ifnet} extracts multiscale features from continuous point locations, which is able to perform upsampling via a learned implicit decoder. However, it has memory and resolution issue due to the involved voxelization process similar to the voxel or implicit field methods. Recently, Dis-PU~\cite{li2021dispu} proposes to disentangle the upsampling subtasks including distribution uniformity and proximity-to-surface via a dense generator and a spatial refiner, and establish a new state-of-the-art. However, all the existing upsampling methods require the input low-res point clouds are clean and noise-free, which prevent them to be used in point completion tasks. 
\section{Method}
\label{sec:methods}

As illustrated in Figure~\ref{fig:pipeline}, there are two phases in our framework, which are low-res point cloud recovery and patch-wise noise-aware upsampling. In the following, Section \ref{sec:sparse_completion} to \ref{sec:iterative_refinement} cover the details of the low-res point cloud recovery, Section \ref{sec:noise_aware_upsampling} introduces the patch-wise noise-aware upsampling, Section \ref{sec:preserving_input} and \ref{sec:all_losses} explain the input information preserving during inference and the overall optimization term in training.

\subsection{Sparse Completion}
\label{sec:sparse_completion}
Given a partial point cloud $\mathcal{P} \in \mathbb{R}^{3}$, our goal is to recover a complete and dense point cloud $\hat{\mathcal{C}} \in \mathbb{R}^{3}$ as close as possible to the ground-truth complete point cloud $\mathcal{C} \in \mathbb{R}^{3}$. Directly regressing or decoding a complete and dense point cloud is not trivial, especially considering many points in $\mathcal{C}$ are missing in $\mathcal{P}$. Instead, we choose to complete the overall structure and shape first. In particular, we regress a complete and sparse point cloud $\mathcal{S}$ providing rough geometry structure with low spatial resolution. The structure of $\mathcal{S}$ should be close to the ground-truth $\mathcal{C}$ as it serves as the foundation of the follow-up steps. Here we adopt PointNet \cite{qi2017pointnet} to extract the features for a partial point cloud, and $\mathcal{S}$ is regressed with five fully connected layers.

\subsection{Symmetry Detection}
\label{sec:symmetry_detection}
Since a considerable part of man-made 3D models share symmetrical characteristic, it is expected that point cloud completion can benefit from symmetry. Although previous works \cite{sung2015data,Wang_2020_CVPR} explore to use symmetry with either supervision or a single pre-defined symmetry plane to generate more reliable and trustworthy points from $\mathcal{P}$, we develop a self-supervised network called, SymNet, to detect planar symmetry automatically. Different from \cite{Gao2020PRSNetPR} that takes dense voxelized volume of fine mesh model as input, we use point cloud $\mathcal{S}$ as it is less memory-consuming and does not require additional complicated voxelization.
Concretely, since $\mathcal{S}$ already has the overall shape, from which we can detect planar symmetry, and then produce additional trustworthy points $\mathcal{P^{'}}$ by mirroring the points from $\mathcal{P}$ based on the predicted plane. 

\begin{figure}[t]
	\centering
	\begin{overpic}[width=0.9\linewidth]{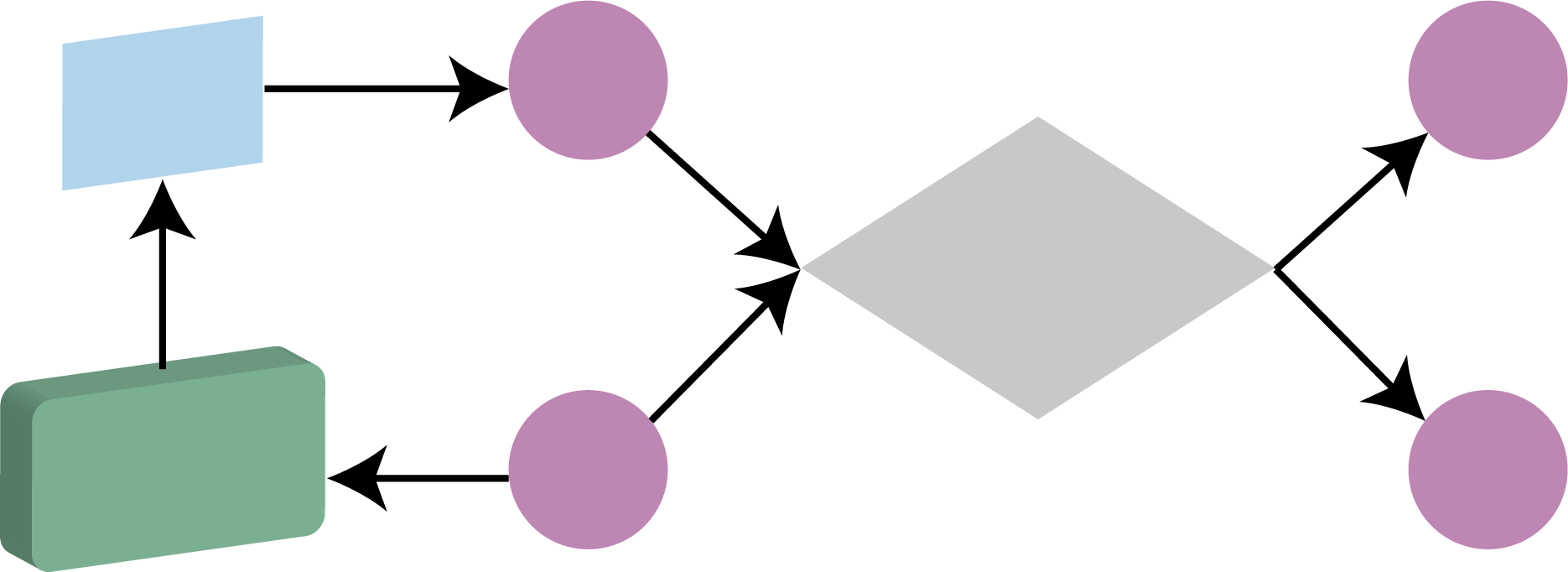}
		\put (36, 5.1) {$\mathcal{S}$}
		\put (36, 29.5) {$\mathcal{Q}$}
		\put (93, 5.1) {$\mathcal{P}$}
		\put (93, 29.5) {$\mathcal{P}^{'}$}
		\put (4.5, 5) {\small SymNet}
		\put (6.8, 28) {$\mathbf{SP}$}
		\put (55.8, 18) {$\mathcal{L}_{sym}\textgreater~\tau~?$}
		\put (82, 23) {Y}
		\put (82, 11) {N}
	\end{overpic}
	\caption{Symmetry Detection. Taking $\mathcal{S}$ as input of SymNet, a plane of symmetry $\mathbf{SP}$ is predicted. $Q$ is produced by applying plane symmetry to $\mathcal{S}$ w.r.t. $\mathbf{SP}$. If $\mathcal{L}_{sym}(\mathcal{Q}, \mathcal{S})$ is greater than a threshold $\tau$, we classify the shape being asymmetrical, and the symmetrical points $\mathcal{P^{'}}$ will not be used and be replaced with $\mathcal{P}$.}
	\vspace{-1mm}
	\label{fig:symmetry}
\end{figure}
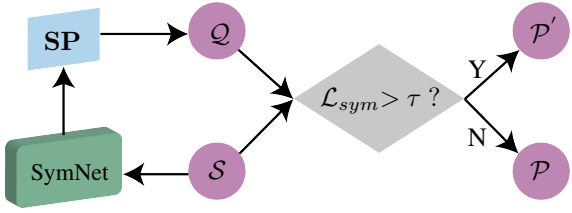

To be specific, as shown in Figure~\ref{fig:symmetry}, for the sparse point cloud $\mathcal{S}$, we adopt PointNet~\cite{qi2017pointnet} as SymNet's feature extractor to get its global representation, and 3 fully connected layers as symmetry plane predictor, which will predict a symmetry plane $\mathbf{SP} = (\mathbf{n}, d)$ in the implicit form, where $\mathbf{n} = (a, b, c)$ is the normal vector of this plane, and the reflection plane is defined as $ax + by + cz + d = 0$. To optimize $\mathbf{SP}$ to be the potential symmetry plane of $\mathcal{S}$, we create a point cloud $Q$ by applying plannar symmetry to $\mathcal{S}$ w.r.t. $\mathbf{SP}$. Each point $\mathbf{q}$ in $Q$ is transformed from the corresponding point $\mathbf{p}$ in $\mathcal{S}$ as:
\begin{equation}
\setlength\abovedisplayskip{4pt}
\setlength\belowdisplayskip{4pt}
\mathbf{q} = \mathbf{p} - 2\frac{\mathbf{p} \cdot \mathbf{n} + d}{\left\lVert \mathbf{n} \right\rVert^2} \mathbf{n}
\end{equation}

After the symmetry plane $\mathbf{SP}$ is available, we get the symmetrical points $\mathcal{P^{'}}$ by applying plane symmetry to $\mathcal{P}$ w.r.t. $\mathbf{SP}$ the same as $\mathcal{Q}$ to $\mathcal{S}$, in order to enrich information from $\mathcal{P}$. Even many man-made 3D models are symmetrical, there are still some asymmetrical ones, e.g., some sectional sofas or lamps in ShapeNet \cite{chang2015shapenet}. Therefore, we add a further step to take care of the asymmetrical models. In particular, if $\mathcal{L}_{sym}(\mathcal{Q}, \mathcal{S})$ is greater than a threshold $\tau$, we classify the shape being asymmetrical, and the symmetrical points $\mathcal{P^{'}}$ will not be used and be replaced with $\mathcal{P}$.

\subsection{Iterative Geometry Refinement}
\label{sec:iterative_refinement}
\begin{figure}[t]
\centering
    \begin{overpic}[width=0.95\linewidth]{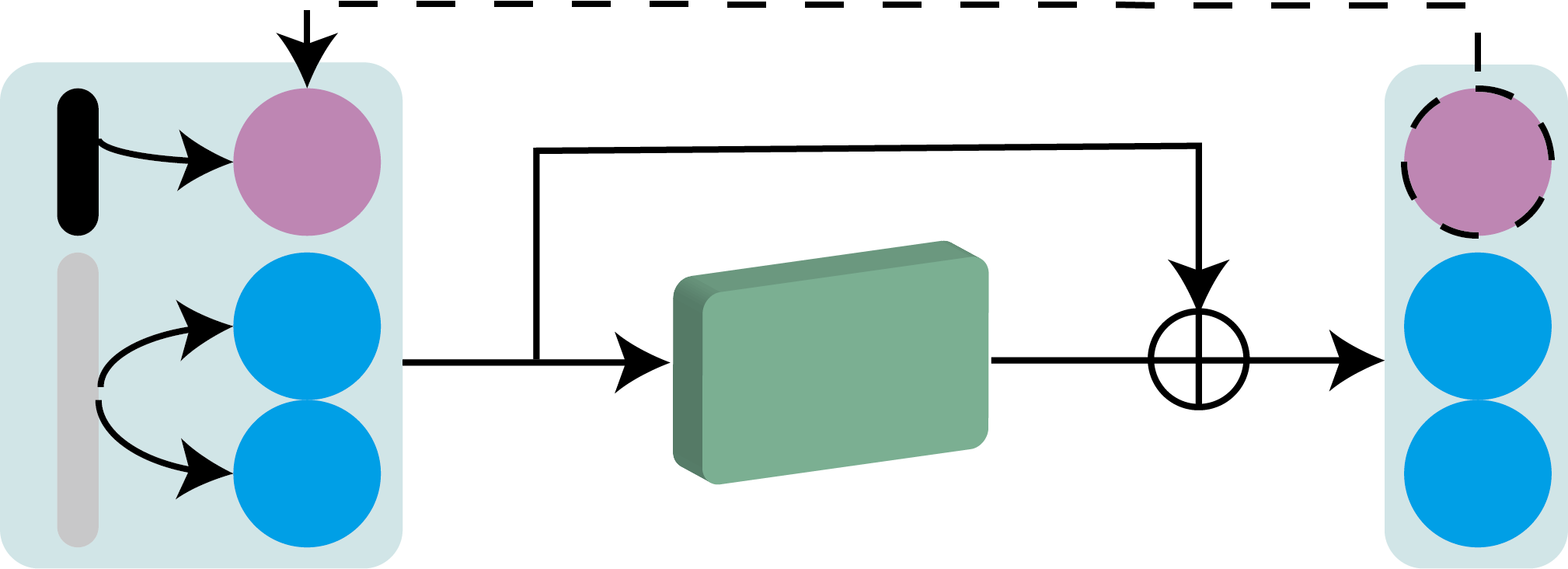}
     \put (18, 25) {\footnotesize $\mathcal{S}$}
     \put (18, 14.5) {\footnotesize $\mathcal{P}$}
     \put (18, 4.75) {\footnotesize $\mathcal{P}^{'}$}
     \put (93, 25) {\footnotesize $\mathcal{S}^{'}$}
     \put (93, 14.5) {\footnotesize $\mathcal{P}$}
     \put (93, 4.75) {\footnotesize $\mathcal{P}^{'}$}
     \put (43, 32.5) {\footnotesize Iterative Refinement}
     \put (48, 12) {\small ResNet}
	 \put (1, 25) {\small 0}
	 \put (1, 10) {\small 1}    
    \end{overpic}
\caption{Iterative ResNet refinement module. We treat predicted $\mathcal{S}$, input partial $\mathcal{P}$ along with its symmetrical points $\mathcal{P}^{'}$ as two parts, attach each part with pseudo labels, feed which into our geometry refinement network ResNet to get $\mathcal{S}^{'}$. And $\mathcal{S}$ will be replaced with refined $\mathcal{S}^{'}$ in an iterative fashion.}
\label{fig:resnet}
\vspace{-1mm}
\end{figure}
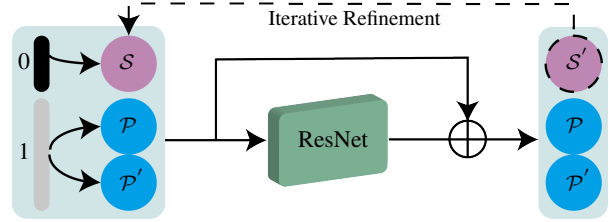

Although the symmetrization process enriches $\mathcal{S}$ by mirroring the input partial points based on the detected symmetry plane, $\mathcal{S}$ is still noisy as accurately recovering from partial shape is not easy. We thus develop an iterative residual refinement strategy to pull points in $\mathcal{S}$ to locate on the underlying surface, further enhancing the structure of $\mathcal{S}$. Specifically, we concatenate $\mathcal{P}, \mathcal{P^{'}}$ and $\mathcal{S}$ along with their pseudo labels, unlike \cite{li2019pu,liu2020morphing} using complicated sampling procedure, we propose to do random down-sampling here. Later, the concatenated features are fed into the geometry refinement network, ResNet, to regress a spatial residual for each point and make $\mathcal{S^{'}}$ finer. Figure~\ref{fig:resnet} shows the whole iterative refinement module. 

Thanks to the effective geometry refinement network and rich symmetry points information enhancement, by replacing the original $\mathcal{S}$ with the refined $\mathcal{S^{'}}$, and repeating the refinement step for a few iterations, a sparse and complete point cloud with higher-fidelity $\Tilde{\mathcal{S}}$ is obtained.

\subsection{Noise-Aware Upsampling}
\label{sec:noise_aware_upsampling}
Existing upsampling~\cite{yang2018foldingnet, yu2018pu, li2019pu} methods focus on densifying clean sparse point clouds with little noise. Though iteratively refined $\Tilde{S}$ is already well structured and with high-fidelity, the noisy points are inevitably introduced from $\mathcal{S}$. Unlike those directly upsample the whole point cloud~\cite{yuan2018pcn, Wang_2020_CVPR}, we focus on local patch to learn finer local pattern taking noise into consideration. We take \cite{li2021dispu} which expands patch points to a denser one in feature space with global and local refinement network, as our upsampling backbone network. Finally, the dense and complete point cloud $\Tilde{\mathcal{C}}$ is obtained by combining the upsampled patches.

\begin{figure}[t]
\includegraphics[width=\linewidth]{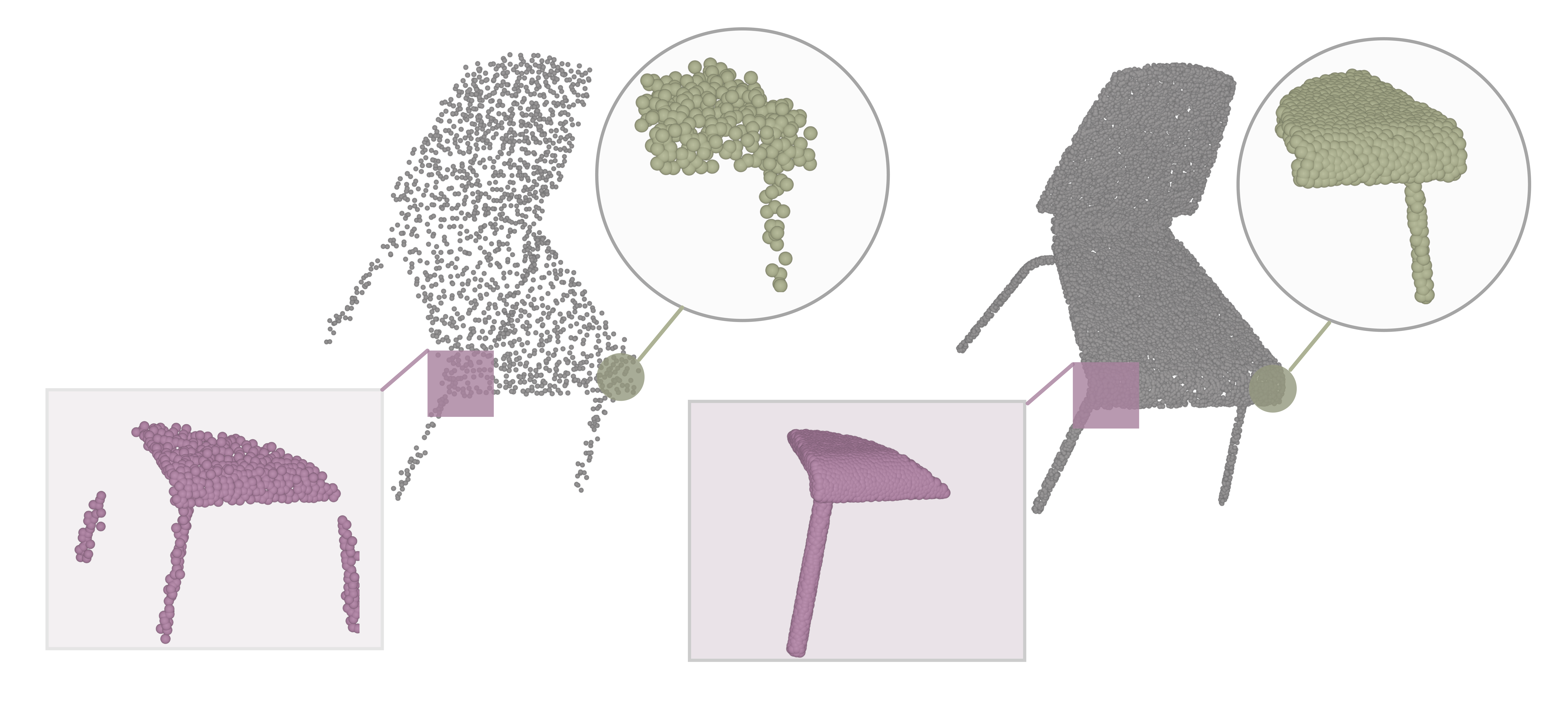}
\caption{Example of the training patch pairs extracted by ball query (yellow) and KNN (purple) strategies from the refined sparse and complete point cloud $\Tilde{\mathcal{S}}$ and its corresponding ground-truth dense and complete point cloud $\mathcal{C}$.}
\label{fig:patch}
\vspace{-1mm}
\end{figure}

\noindent \textbf{Patch Extraction.} The state-of-the-art point cloud upsampling method extracts training patches by cropping overlapped patches by geodesic distance from triangle mesh and afterwards apply uniform point sampling~\cite{li2021dispu}. As no associated triangle mesh available during testing, k-nearest neighbors (KNN) is adopted to extract a local patch around per farthest sampled seed~\cite{li2021dispu}. In our case, even in training the triangle mesh corresponding to the obtained sparse and complete point cloud $\Tilde{\mathcal{S}}$ is not available. Thus, KNN has to be used in extracting the training patches if we directly applied the upsampling method~\cite{li2021dispu} to completion task. However, this is problematic because applying KNN to extract fixed number of points between $\Tilde{\mathcal{S}}$ and its corresponding ground-truth dense point cloud $\mathcal{C}$ could result in huge mismatch between the extract patches as shown in Figure~\ref{fig:patch}. 

Therefore, we propose to generate a training pair in a ball query fashion to make sure the patch pair overlap with each other well. To be specific, each patch $\mathcal{A}(i)$ around point $c_{i}$ in $\Tilde{S}$ is defined as $\mathcal{A}(i) = \{x \mid x \in \Tilde{S}, \Vert x - c_{i} \Vert \leq r\}$ and $\vert \mathcal{A}(i) \vert = \mathrm{K}$, where $c_i$ is a furthest point sampling seed of $\mathcal{C}$ in training and $\Tilde{S}$ in testing. 
To demonstrate the effectiveness of the proposed patch pair extraction strategy, we compare the two patch extraction approaches: KNN vs ball query, as show in Figure~\ref{fig:patch}. KNN strategy cannot produce reasonable patch pairs since the density of input sparse point cloud is not uniform and noisy; on the contrary, our proposed ball query can generate meaningful patch pairs that cover the same region of points between the sparse and its corresponding ground-truth point clouds.

\noindent \textbf{Outlier Removal.} To reduce the impact from noisy points in $\Tilde{S}$, we apply a quite simple but effective radius outlier filtering before feeding into the upsampling network. Specifically, for each point $x \in \Tilde{S}$, we define its neighbors as $\mathcal{N}_{x} = \{y \mid y \in \Tilde{S}, \Vert x-y \Vert < r \}$, whose cardinality is represented as $\vert \mathcal{N}_{x} \vert$. If $\vert \mathcal{N}_{x} \vert < \gamma$, where $\gamma$ is a threshold, we classify $x$ as an outlier and remove it. In the same way, we also apply outlier filtering on the dense and completed point cloud $\Tilde{\mathcal{C}}$. 

\subsection{Preserving Input Information.}
\label{sec:preserving_input}
During inference, to further preserve the details and trustworthy points from its input partial point cloud, with the help of the symmetry detection module, we merge the input points $\mathcal{P}$, its symmetrical points $\mathcal{P}^{'}$, and $\Tilde{\mathcal{C}}$ to get a more reliable $\Tilde{\mathcal{C}} = \mathbf{fps}(\mathcal{P} \cup \mathcal{P}^{'} \cup \Tilde{\mathcal{C}}, \vert \Tilde{\mathcal{C}} \vert)$, where $\mathbf{fps}(\mathcal{X}, \mathbf{k})$ means applying farthest point sampling  \cite{qi2017pointnet++} on $\mathcal{X}$ to obtain $\mathbf{k}$ points, $\vert \mathcal{C} \vert$ is the cardinality the ground-truth point cloud.

\subsection{Optimization}
\label{sec:all_losses}
In the training process, taking the given partial point cloud $\mathcal{P}$ as input and its corresponding ground-truth complete point cloud $\hat{\mathcal{C}}$, our completion network produces sparse but complete $\mathcal{S}$, its symmetrical points $\mathcal{Q}$ as well as refined points $\mathcal{\Tilde{S}}$ and the upsampled $\Tilde{\mathcal{C}}$. We define the objective functions for each part as follows:
\begin{equation}
    \mathcal{L}_{sparse} = \mathrm{CD}_{1}(\mathcal{S}, \hat{\mathcal{S}}) + \mathrm{EMD}(\mathcal{S}, \hat{\mathcal{S}})
\end{equation}
\begin{equation}
    \mathcal{L}_{sym} = \mathrm{CD}_{2}(\mathcal{Q}, \hat{\mathcal{S}})
\end{equation}
\begin{equation}
    \mathcal{L}_{refine} = \mathrm{CD}_{1}(\Tilde{\mathcal{S}}, \hat{\mathcal{S}})
\end{equation}
\begin{equation}
    \mathcal{L}_{up} = \mathrm{CD}_{1}(\Tilde{\mathcal{C}}, \hat{\mathcal{C}})
\end{equation}

\noindent where $\mathrm{CD}_{1}$ and $\mathrm{CD}_{2}$ denote the Chamfer Distance with L1 and L2 norm respectively, and $\mathrm{EMD}$ is the Earth Mover Distance. The overall optimization term $\mathcal{L}$ is formulated as: 
\begin{equation}
    \mathcal{L} = \mathcal{L}_{sparse} + \mathcal{L}_{sym} + \mathcal{L}_{refine} + \mathcal{L}_{up}
\end{equation}
\section{Experiments}
\label{sec:experiments}

\subsection{Datasets}
\textbf{ShapeNet}.
We train and validate our method on the dataset derived from PCN~\cite{yuan2018pcn}, containing synthetic CAD models created from a subset of ShapeNet~\cite{chang2015shapenet}, which composed of 8 specific categories and 30974 models in total. For each 3D model, the complete point cloud as ground truth is uniformly sampled from its associated triangle mesh, and the  corresponding partial point cloud is produced by back-projecting a depth map rendered from a random viewpoint.

\textbf{KITTI}.
KITTI is a real-world dataset~\cite{geiger2013vision} scanned by LIDAR. In order to validate the generalization capability of our method, for each frame, point clouds are extracted from the bounding boxes labeled as car, which results in 2483 partial point clouds. Since there exists a large variation in number of points of these point clouds, we randomly drop or duplicate 2048 points for each partial point cloud, feeding which to our framework.

\begin{table*}[htbp]
  \centering
    \begin{tabular}{c|c|c|c|c|c|c|c|c|c}
        \hline
        \multirow{2}{*}{Method} & \multicolumn{9}{c}{Average Mean Chamfer Distance ($10^{3}$)} \\
          \cline{2-10}
          \multicolumn{1}{c|}{}
          & \multicolumn{1}{c}{Average} & \multicolumn{1}{c}{Airplane} & \multicolumn{1}{c}{Cabinet} & \multicolumn{1}{c}{Car} & \multicolumn{1}{c}{Chair} & \multicolumn{1}{c}{Lamp} & \multicolumn{1}{c}{Sofa} & \multicolumn{1}{c}{Table} & \multicolumn{1}{c}{Vessel} \\
          \hline\hline
    3D-EPN~\cite{dai2017shape} & 20.15 & 13.16 & 21.80 & 20.31 & 18.81 & 25.75 & 21.09 & 21.72 & 18.54 \\
    FC~\cite{yuan2018pcn} & 9.80 & 5.70 & 11.02 & 8.78 & 10.97 & 11.13 & 11.76 & 9.32  & 9.72 \\
    Folding~\cite{yuan2018pcn} & 10.07 & 5.97 & 10.83 & 9.27 & 11.25 & 12.17 & 11.63 & 9.45 & 10.03 \\
    PN2~\cite{yuan2018pcn}  & 14.00 & 10.30  & 14.74 & 12.19 & 15.78 & 17.62 & 16.18 & 11.68 & 13.52 \\
    PCN-CD~\cite{yuan2018pcn} & 9.64 & 5.50 & 10.63 & 8.70 & 11.00 & 11.34 & 11.67 & 8.59  & 9.67 \\
    PCN-EMD~\cite{yuan2018pcn} & 10.02 & 5.85 & 10.69 & 9.08  & 11.58 & 11.96 & 12.21 & 9.01 & 9.79 \\
    TopNet~\cite{tchapmi2019topnet} & 9.89 & 6.24 & 11.63 & 9.83 & 11.50 & 9.37 & 12.35 & 9.36 & 8.85 \\
    MSN~\cite{liu2020morphing} * & 9.97 & 5.59 & 11.95 & 10.74 & 10.63 & 10.75 & 11.88 & 8.72 & 9.49 \\
    GRNet~\cite{xie2020grnet} & 9.03 & 6.41 & 10.91 & 9.63 & 9.64 & 7.97 & 10.77 & 8.77 & 8.11 \\
    CRN~\cite{Wang_2020_CVPR} & 8.51 & 4.79 & 9.97 & \textbf{8.31} & 9.49 & 8.94 & 10.69 & 7.81 & 8.05 \\
    NSFA~\cite{zhang2020detail} & 8.06 & 4.76 & 10.18 & 8.63 & 8.53 & \textbf{7.03} & 10.53 & \textbf{7.35} & 7.48 \\
    PMP-Net~\cite{Wen_2021_CVPR} & 8.66 & 5.50 & 11.10 & 9.62 & 9.47 & 6.89 & 10.74 & 8.77 & 7.19 \\ \hline
    Ours & \textbf{7.90} & \textbf{4.48} & \textbf{9.55} & 8.58 & \textbf{8.28} & 8.03 & \textbf{9.69} & 7.46 & \textbf{7.14} \\ \hline
    \end{tabular}%

  \caption{Quantitative comparison results on ShapeNet for resolution 16384 evaluated with L1 Chamfer Distance $\times 10^3$. Best in \textbf{bold}. * denotes extra data is used in training. Our method achieves the best performance on average from 5 out of 8 categories.}
  \label{tab:shapenet}%
\end{table*}%

\begin{figure*}[ht]
	\centering
	
	\begin{overpic}[width=\linewidth]{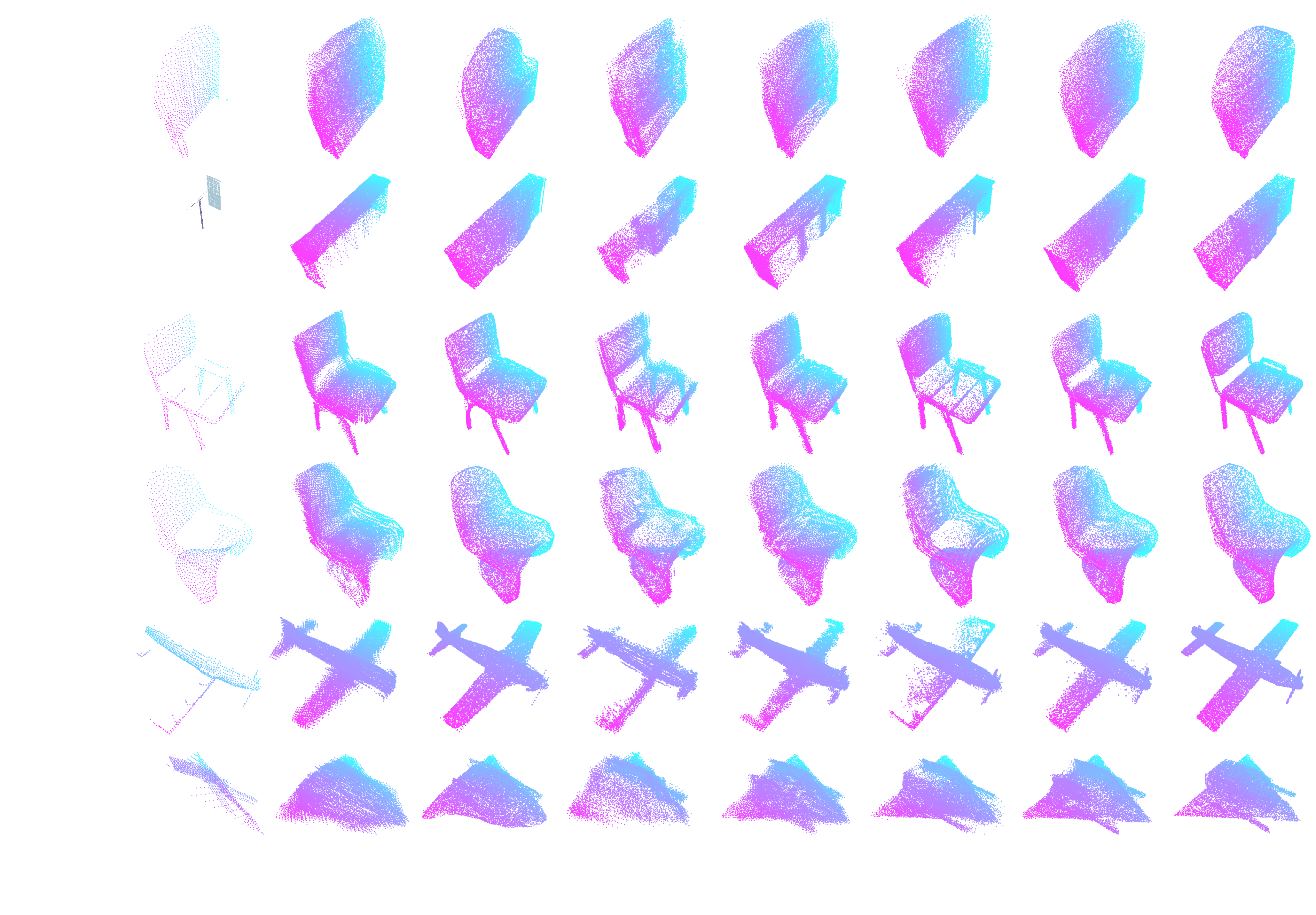}
		\put (4, 62) {(a)}
		\put (4, 51) {(b)}
		\put (4, 39) {(c)}
		\put (4, 28) {(d)}
		\put (4, 19) {(e)}
		\put (4, 8) {(f)}
		
		\put (14, 3) {\small Input}
		\put (25, 3) {\small PCN}
		\put (35, 3) {\small MSN}
		\put (46, 3) {\small GRNet}
		\put (58, 3) {\small CRN}
		\put (70, 3) {\small NSFA}
		\put (82, 3) {\small Ours}
		\put (94, 3) {\small GT}
	\end{overpic}
	\vspace{-8mm}
	\caption{Completion results on ShapeNet. Each column from left to right is: input partial point cloud, results of PCN~\cite{yuan2018pcn},
		MSN~\cite{liu2020morphing}
		GRNet~\cite{xie2020grnet}, CRN~\cite{Wang_2020_CVPR}, NSFA~\cite{zhang2020detail}, ours, and the ground truth (GT). It is observed that our results show better visual similarity to the ground-truth point clouds compared with the state-of-the-art point cloud completion methods.}
	\label{fig:new_shapenet}
	\vspace{-3mm}
\end{figure*}
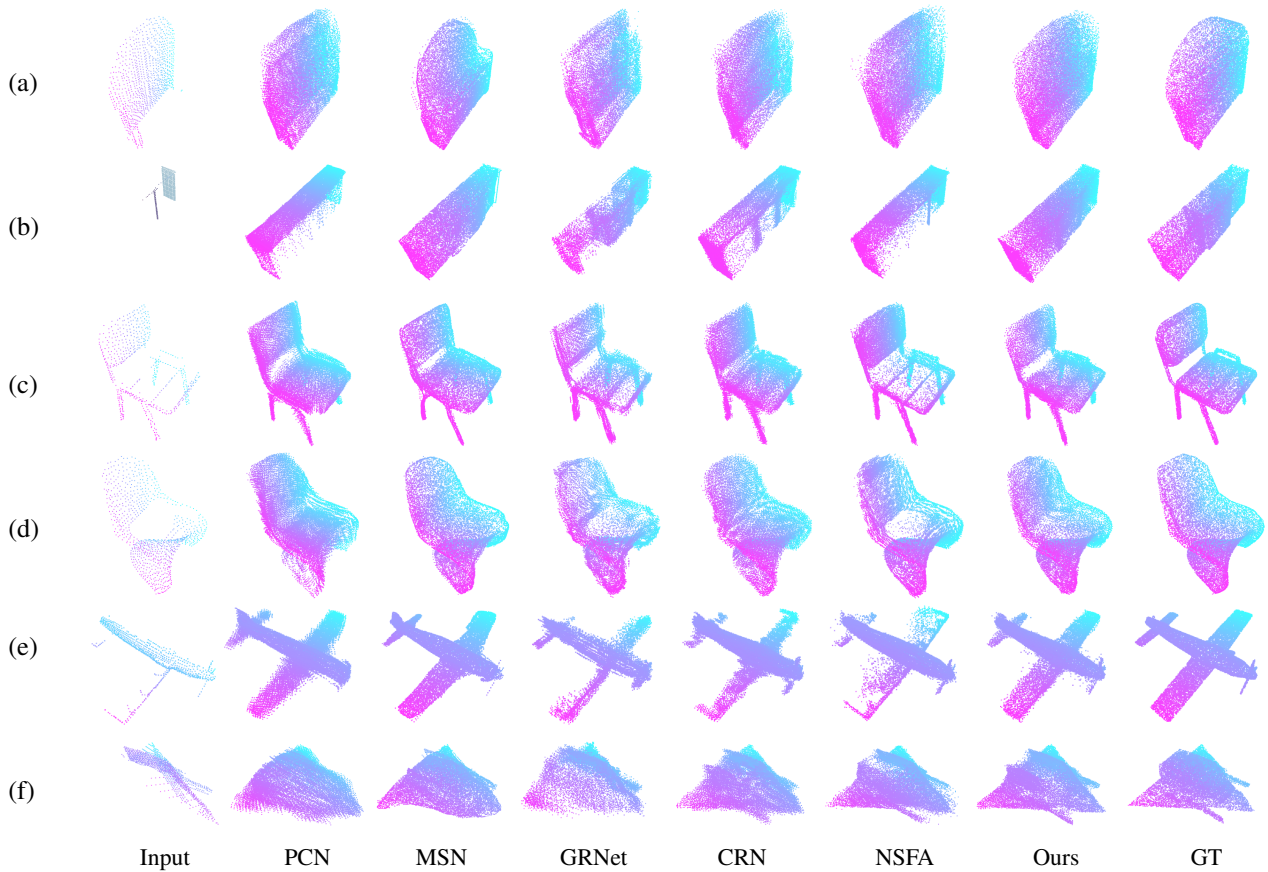

\subsection{Metrics}
We use bi-directional Chamfer Distance (CD), and F-Score as metrics to evaluate the accuracy of completed point clouds on ShapeNet following~\cite{fan2017point,yuan2018pcn,Wang_2020_CVPR,liu2020morphing,xie2020grnet}. In particular, Chamfer Distance with L1 norm is used in the following quantitative evaluations. 

To perform evaluations on KITTI dataset where ground-truth is not available, we use Fidelity, Minimal Matching Distance (MMD), and Consistency following~\cite{yuan2018pcn}. Fidelity is the average distance between the input partial point cloud and its completed result, indicating how well the input information is preserved. MMD is the lowest Chamfer Distance between completed point cloud and the car point clouds from ShapeNet, which measures how much it resembles a typical car. And Consistency is the average Chamfer Distance between the completed point clouds of the same car in consecutive frames.
\begin{table}[thbp]
\centering
\begin{tabular}{l|cccc|c}
\hline
\multicolumn{1}{c|}{\multirow{2}{*}{Category}} & \multicolumn{5}{c}{Method}\\
\cline{2-6}
\multicolumn{1}{c|}{}& PCN & CRN & GRNet & NSFA & Ours  \\ \hline\hline
Airplane             &   0.88  &  0.90   &   0.84    &   0.91   &   \textbf{0.93}    \\
Cabinet              &   0.65  &  0.57   &   0.62    &   0.66   &   \textbf{0.70}    \\
Car                  &   0.73  &  0.68   &   0.68    &   0.72   &   \textbf{0.73}    \\
Chair                &   0.63  &  0.62   &   0.67    &   0.74   &   \textbf{0.75}    \\
Lamp                 &   0.64  &  0.67   &   0.76    &   \textbf{0.82}   &   0.78    \\
Sofa                 &   0.58  &  0.54   &   0.61    &   0.63   &   \textbf{0.66}    \\
Table                &   0.77  &  0.71   &   0.75    &   \textbf{0.83}   &   0.81    \\
Vessl                &   0.70  &  0.74   &   0.75    &   0.79   &   \textbf{0.81}    \\ 
\hline
Average              &   0.70  &  0.68   &   0.71    &   0.76   &   \textbf{0.77}    \\ \hline
\end{tabular}
\vspace{1mm}
\caption{Quantitative comparison results on ShapeNet with F-Score@1\%$\uparrow$. Best in \textbf{bold}.}
\label{tab:fscore}
\end{table}

\begin{table*}[thbp]
  \centering
    \begin{tabular}{c|c|c|c|c|c|c|c|c|c}
        \hline
        \multirow{2}{*}{Model} & \multicolumn{9}{c}{Average Mean Chamfer Distance ($10^{4}$)} \\
          \cline{2-10}
          \multicolumn{1}{c|}{}
          & \multicolumn{1}{c}{Average} & \multicolumn{1}{c}{Airplane} & \multicolumn{1}{c}{Cabinet} & \multicolumn{1}{c}{Car} & \multicolumn{1}{c}{Chair} & \multicolumn{1}{c}{Lamp} & \multicolumn{1}{c}{Sofa} & \multicolumn{1}{c}{Table} & \multicolumn{1}{c}{Vessel} \\
          \hline\hline
    Folding~\cite{yuan2018pcn} & 7.14 & 3.15 & 7.94 & 4.68 & 9.23 & 9.23 & 8.90 & 6.69 & 7.33 \\
    AtlasNet~\cite{atlasnet} & 4.52 & 1.75 & 5.10 & 3.24 & 5.23 & 6.34 & 5.99 & 4.36 & 4.18 \\
    TopNet~\cite{tchapmi2019topnet} & 5.15 & 2.15 & 5.62 & 3.51 & 6.35 & 7.50 & 6.95 & 4.78 & 4.36 \\
    MSN~\cite{liu2020morphing} * & 4.76 & 1.54 & 7.25 & 4.71 & 4.54 & 6.48 & 5.89 & 3.80 & 3.85 \\
    NSFA~\cite{zhang2020detail} & 4.28 & 1.75 & 5.31 & 3.43 & 5.01 & 4.73 & 6.41 & 4.00 & 3.56 \\
    CRN~\cite{Wang_2020_CVPR} & 3.75 & 1.46 & 4.21 & 2.97 & 3.24 & 5.16 & 5.01 & 3.99 & 3.96 \\
    PCN~\cite{yuan2018pcn} & 4.02 & 1.40 & 4.45 & 2.45 & 4.84 & 6.24 & 5.13 & 3.57 & 4.06 \\
    SoftPoolNet~\cite{softpool} & 5.94 & 4.01 & 6.23 & 5.94 & 6.81 & 7.03 & 6.99 & 4.84 & 5.70 \\
    PF-Net~\cite{huang2020pf} & 3.80 & 1.55 & 4.43 & 3.12 & 3.96 & 4.21 & 5.87 & 3.35 & 3.89 \\
    GRNet~\cite{xie2020grnet} & 2.72 & 1.53 & 3.62 & 2.75 & 2.95 & 2.65 & 3.61 & 2.55 & 2.12 \\
    ASHF-Net~\cite{zong2021ashf} & 2.56 & 1.40 & 3.49 & 2.32 & 2.82 & \textbf{2.52} & 3.48 & 2.42 & \textbf{1.99} \\ \hline
    Ours & \textbf{2.42} & \textbf{0.87} & \textbf{3.02} & \textbf{2.25} & \textbf{2.86} & 2.92 & \textbf{3.32} & \textbf{1.95} & 2.18 \\ \hline
    \end{tabular}%
  \vspace{0.5mm}
  \caption{Quantitative comparison results on ShapeNet for resolution 16384 evaluated with L2 Chamfer Distance $\times 10^4$. Best in \textbf{bold}. * denotes extra data is used in training.}
  \label{tab:shapenet2}%
\end{table*}%
\subsection{Implementation Details}
We implement our network with the deep learning framework PyTorch~\cite{pytorch} along with PyTorch Lightning~\cite{pytorchlightning}. For the patch-wise upsampling network, we split the whole input points into 24 patches in training and 16 in testing. The Adam optimizer with initial learning rate of 0.001 is used during training process, and the learning rate is linearly decayed per 3k steps by a rate of 0.8. The number of iterations in our iterative geometry refinement is set to 2. Please refer to the supplementary material for details. 

\subsection{Evaluations on ShapeNet Dataset}

We compare our approach with the state-of-the-art (SOTA) methods on ShapeNet dataset. The quantitative comparison of point cloud completion performance is shown in Table~\ref{tab:shapenet} evaluated with $\mathrm{CD}_{1}$ (lower is better) and, F-Score@1\% (higher is better) in Table~\ref{tab:fscore}. For Chamfer Distance, our method achieves the best performance on average from 5 out of 8 categories. The result of CRN~\cite{Wang_2020_CVPR} is from its original paper, and we follow GRNet\cite{xie2020grnet} to produce the result of MSN~\cite{liu2020morphing} at the resolution of 16384. Our method has the highest value under the metric of F-Score@1\% as well.

As performance on ShapeNet with L2 Chamfer Distance ($\mathrm{CD}_{2}$) is reported in recent works \cite{huang2020pf,softpool,xie2020grnet}, we also perform additional quantitative evaluations with L2 Chamfer Distance ($\mathrm{CD}_{2}$) for comprehensive comparisons as shown in Table~\ref{tab:shapenet2}. In particular, the values of CRN \cite{Wang_2020_CVPR} and PF-Net \cite{huang2020pf} are from ASHF-Net~\cite{zong2021ashf}, the numbers of other methods are from their papers. From the table, we observe that our method again outperforms the existing methods similar to the L1 Chamfer Distance ($\mathrm{CD}_{1}$) evaluation in Table~\ref{tab:shapenet}, which confirms the superior performance of our method on both L1 and L2 Chamfer Distance evaluations.

Qualitative comparisons are shown in Figure~\ref{fig:new_shapenet}. We observe that our results show better visual similarity compared with the ground-truth point clouds, where missing structures are more accurately recovered (e.g., the recovered half of the cabinet in row (b)) and details are better preserved from the input partial point cloud (e.g., the details of the chairs in row (c) and (d)). The airplane in row (e) is a challenging case where we observe none of the completion results is perfect, but our result is closest to the ground-truth with minor noise compared against other completion results either losing too much detail like PCN and CRN, or adding too much noise like GRNet and NSFA. 

To better understand the how each method performs for heavy occlusions cases, we can take a closer look at the results of the last row (f) in Figure~\ref{fig:new_shapenet}. In particular, the top part of the airplane of the ground truth (GT) is covered, however, due to the cover is completely missing in the input partial point cloud, the completion results of other methods and ours only recover the supporting structure but not the cover itself. Given all the results are not perfect, our completion result best recovers the supporting structure of the cover of the airplane, which is the closest to the GT.

\subsection{Ablation Studies}

\begin{table}[b!]
  \centering
  \resizebox{\linewidth}{!}{
    \begin{tabular}{c|cccccc|c}
    \hline
    Model & SymNet & ResNet & Outlier Removal & Input Preservation & $\mathrm{CD}_1$\\
    \hline \hline
    A & & & & & 11.789 \\
    B &  & \Checkmark & \Checkmark & \Checkmark & 8.566 \\
    C & \Checkmark & & \Checkmark & \Checkmark & 8.460 \\
    D &  &  & \Checkmark & \Checkmark & 8.911 \\
    E & \Checkmark & \Checkmark & & \Checkmark & 8.310 \\
    F & \Checkmark & \Checkmark & \Checkmark & & 8.385 \\
    \hline
    Full & \Checkmark & \Checkmark & \Checkmark & \Checkmark & \textbf{8.284} \\ \hline
    \end{tabular}
    }
    \vspace{1mm}
    \caption{Comparing the performance evaluated with L1 Chamfer Distance $\times 10^3$ for each variance and our full method.}
  \label{tab:ablation}
\end{table}

To validate of the effectiveness of the major modules of our method, we implement six variations of our method for ablation study: (A) a baseline without any of the proposed components such as SymNet, ResNet, outlier removal, and preserving input information; (B) removing the SymNet module only, where the symmetrical points $\mathcal{P}^{'}$ is replaced with $\mathcal{P}$ to keep the cadinality of the refined points $\Tilde{\mathcal{S}}$ constant; (C) removing the iterative ResNet only; (D) removing both the SymNet and iterative ResNet; (E) removing outlier removal only; (F) removing preserving input information. Evaluations of the six variations and our full model with $\mathrm{CD}_{1}$ are provided in Table~\ref{tab:ablation}. All variations are conducted on the Chair category of ShapeNet. It is observed that removing each major module in our framework lowers the performance compared with our full model, which validates the effectiveness of the proposed components. In particular, the baseline model (A) without any of the proposed components performs the worst compared with all other models. 

\begin{figure}[t!]
\centering
    \begin{overpic}[width=0.9\linewidth]{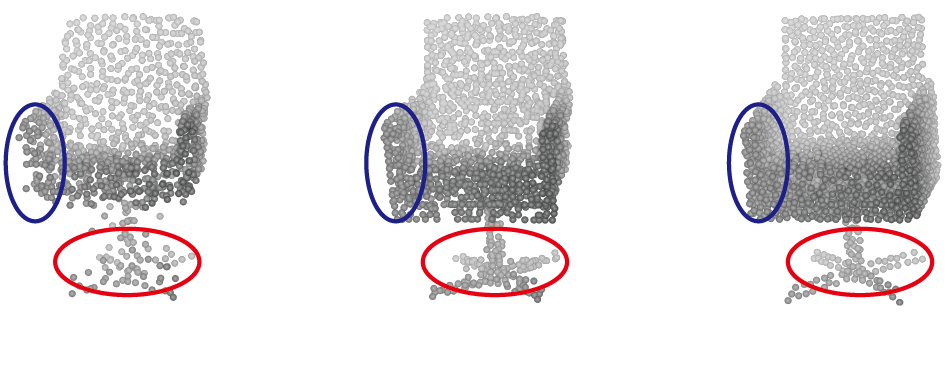}
     \put(8, 3) {before}
     \put(48, 3) {after}
     \put(88, 3) {GT}
    \end{overpic}
\caption{Effects of our iterative geometry refinement module. Left: before iterative geometry refinement ($\mathcal{S}$); Middle: after iterative geometry refinement ($\Tilde{\mathcal{S}}$); Right: Ground Truth.
}
\label{fig:georefine}
\end{figure}

\begin{figure}[b]
\centering
    \begin{overpic}[width=0.9\linewidth]{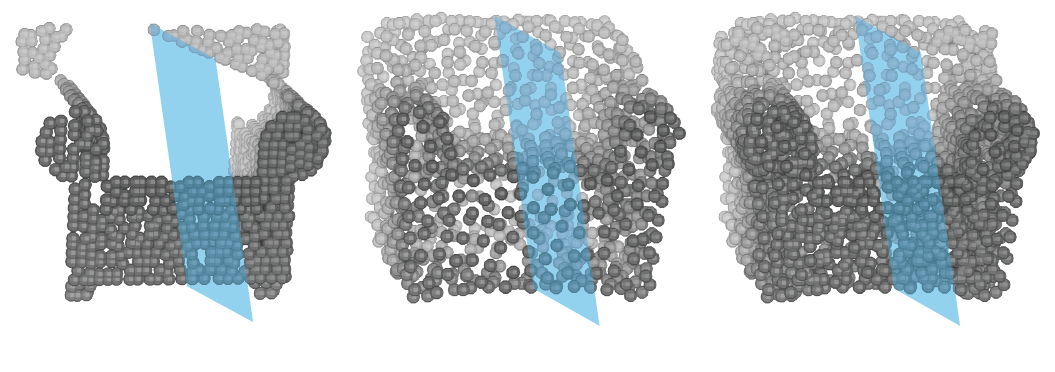}
     \put (15, 2) {(a)}
     \put (47, 2) {(b)}
     \put (83, 2) {(c)}
    \end{overpic}
\caption{Effects of our symmetry detection module. From left to right: (a) input partial point cloud $\mathcal{P}$, (b) sparse completion result $\mathcal{S}$, (c) enhanced point cloud of symmetry detection. The detected symmetry plane $\mathrm{SP}$ overlaid on top of the three point clouds in light blue for reference.
}
\label{fig:sym}
\vspace{-1mm}
\end{figure}

We provide visual examples in the following to help understand the effects of with and without the proposed major modules. Figure~\ref{fig:georefine} shows the completion results of with and without the iterative refinement module. It is clearly observed that, after our iterative refinement, the shape of the armrests and legs (i.e., highlighted in blue and red ellipses) becomes sharper and clearer, which is closer to its corresponding ground truth. Similarly, the visual comparisons of using the symmetry detection module or not are shown in Figure~\ref{fig:sym}. We observe that the armrests in (c) are more complete and in better shape than that in (b), because the input points and their symmetrical counterpart are added based on the symmetry detection.

\begin{figure}[t]
\centering
    \begin{overpic}[width=0.9\linewidth]{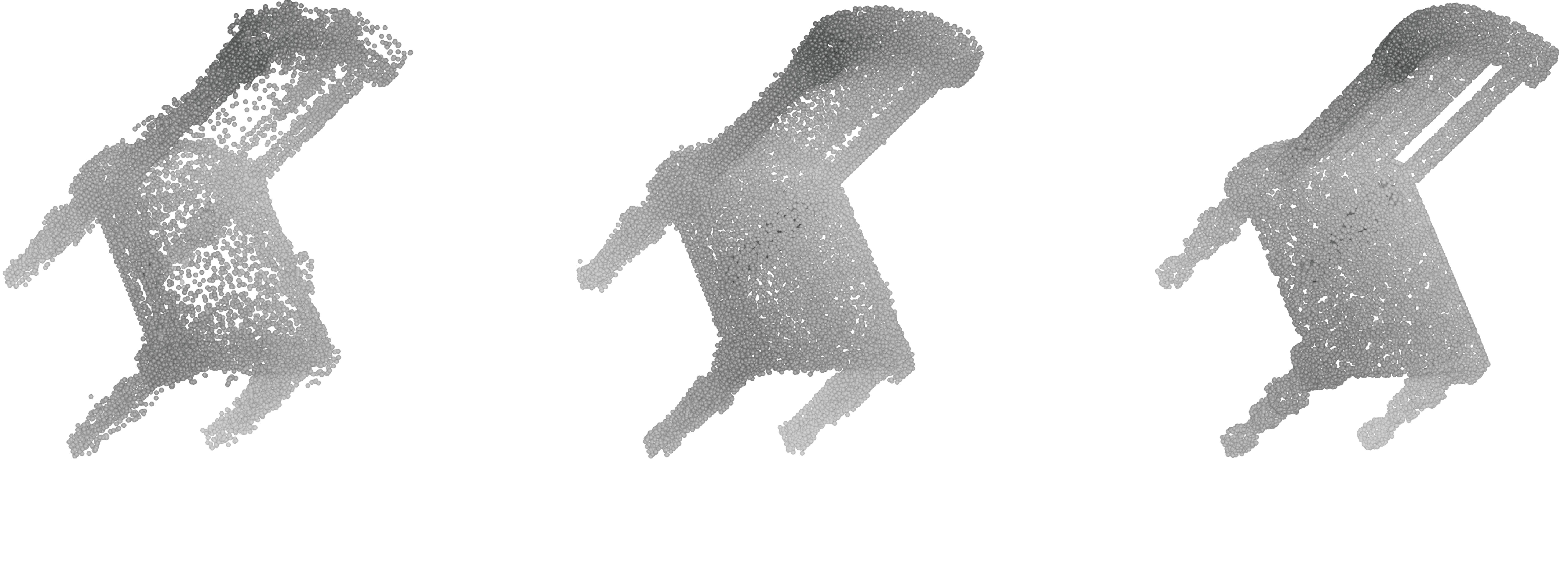}
     \put (12, 2) {KNN}
     \put (47, 2) {Ball query}
     \put (83, 2) {GT}
    \end{overpic}
\caption{Effects of patch extraction strategies. Left: upsampled points from patches extracted by KNN; Middle: upsampled points from patches extracted by ball query; Right: Ground Truth.}
\label{fig:ballknn}
\end{figure}

In the upsampling stage, our patch extraction and outlier removal modules are critical to achieve high-fidelity point cloud completion. As shown in Figure~\ref{fig:patch} in method section, without the proposed patch extraction strategy, no meaningful patch pairs can be extracted for training the patch-wise upsampling network. 
 
\begin{figure}[tb]
\centering
    \begin{overpic}[width=0.9\linewidth]{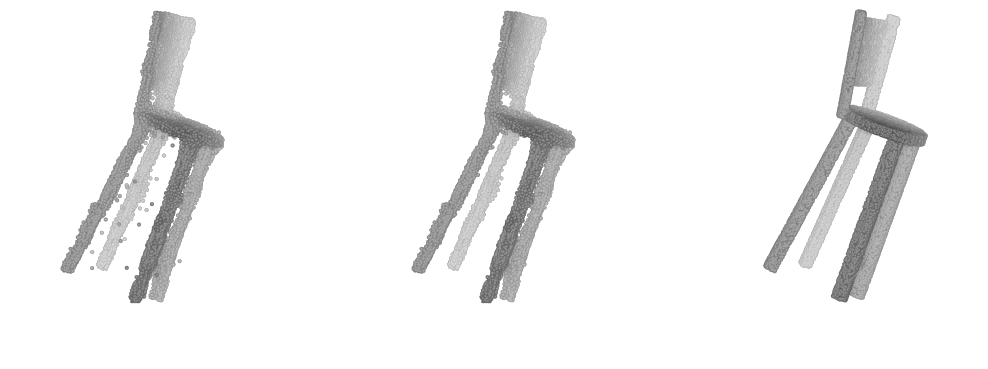}
     \put(8, 2) {\small w/o OR}
     \put(43, 2) {\small with OR}
     \put(84, 2) {\small GT}
    \end{overpic}
\caption{Effects of our outlier removal (OR) module. Left: up-sampled points w/o outlier removal; Middle: up-sampled points with outlier removal; Right: Ground Truth.
}
\label{fig:outlier}
\end{figure}

We provide the comparison of using the two patch extraction approaches to generate training patch pairs in Figure~\ref{fig:ballknn}, it is observed that the upsampled point cloud using our proposed ball query strategy is much better than the KNN based result. As it is inevitable to amplify noise in upsampling, with the help of our simple but effective outlier removal, cleaner upsampled point cloud with less noise and artifacts is obtained as demonstrated in Figure~\ref{fig:outlier}.

As the number of iterations in the proposed iterative refinement module is a hyper-parameter in our method, we measure the performance by varying the number of iterations to validate our choice of the parameter as shown in Table~\ref{tab:resnet}.
From the table, we can see that as the number of iterations of ResNet increases, the performance improvement saturates. Therefore, to balance between the computational cost and accuracy, we fix the number of iterations to 2 for all the experiments.

\begin{table}[t!]
  \centering
  \resizebox{0.9\linewidth}{!}{
    \begin{tabular}{c|ccc}
    \hline
    Iterations of ResNet & 1 & 2 & 3 \\
    \hline
    $\mathrm{CD}_1$ & 8.326 & 8.265 & 8.264 \\ \hline
    \end{tabular}
    }
    \vspace{1mm}
    \caption{Comparing the performance evaluated with L1 Chamfer Distance $\times 10^3$ when changing the number of iterations of our iterative geometry refinement module.}
  \label{tab:resnet}
\end{table}

\begin{figure}[t!]
    \centering
    \begin{overpic}[width=\linewidth]{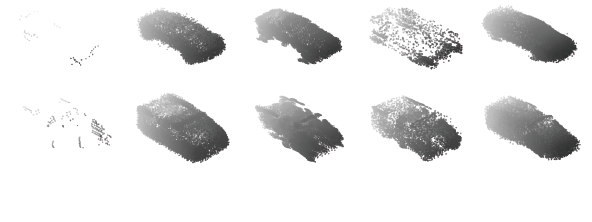}
    \put (26, 22.3) {\footnotesize 0.0092}
    \put (46, 22.3) {\footnotesize 0.0088}
    \put (66, 22.3) {\footnotesize 0.0155}
    \put (86, 22.3) {\footnotesize 0.0042}
    
    \put (26, 6) {\footnotesize 0.0101}
    \put (46, 6) {\footnotesize 0.0090}
    \put (66, 6) {\footnotesize 0.0119}
    \put (86, 6) {\footnotesize 0.0087}

    \put(8, 1) {\footnotesize Input}
    \put(26, 1) {\footnotesize GRNet}
    \put(48, 1) {\footnotesize CRN}
    \put(67, 1) {\footnotesize NSFA}
    \put(88, 1) {\footnotesize Ours}
    \end{overpic}
    \caption{Completion results on KITTI compared with GRNet~\cite{xie2020grnet}, CRN~\cite{Wang_2020_CVPR}, NSFA~\cite{zhang2020detail}. Each value under the completed point cloud is the Fidelity error, which shows our method achieves the lowest Fidelity error among the compared approaches.
    }
    \label{fig:KITTI}
\end{figure}
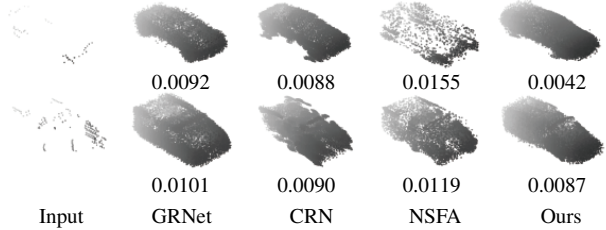

\subsection{Evaluations on KITTI Dataset}

To evaluate the generalization capability of our method, we apply our method on the KITTI car dataset, which is acquired by real-world LiDAR scans. 
As ground-truth point clouds are not available in KITTI car dataset, 
quantitative evaluations with Fidelity, MMD, and Consistency following~\cite{yuan2018pcn} are performed. In particular, we evaluate the completion results of our method compared with that of the state-of-the-art (SOTA) point cloud completion methods including GRNet~\cite{xie2020grnet}, CRN~\cite{Wang_2020_CVPR}, and NSFA~\cite{zhang2020detail}.

As shown in Table~\ref{tab:kitti}, our method demonstrates superior performance in preserving the details of geometry measured by Fidelity, where $41.6\%$ error reduction is achieved compared with GRNet, which is the best performed existing method on this metric. Given GRNet's relatively low MMD and Consistency scores but high Fidelity value, it indicates the completion results of GRNet are of high bias and low variance, which means their results could be constantly over-smoothing and noisy. 

In order to better understand the characteristics of each method, we also provide the examples of visual comparisons against GRNet, CRN, and NSFA as shown in Figure~\ref{fig:KITTI}. It is observed that the results of our method consistently resemble the shape of cars, while the results of other methods are noisy and losing details, which makes their completed point clouds visually unrecognizable if they are cars or not. We can also see that our method preserves the details from the input partial point cloud more accurately than other methods, which confirms with our low error in Fidelity evaluation.

\begin{figure}[t]
	\centering
	\begin{minipage}[c]{0.23\linewidth}
		\includegraphics[width=\linewidth]{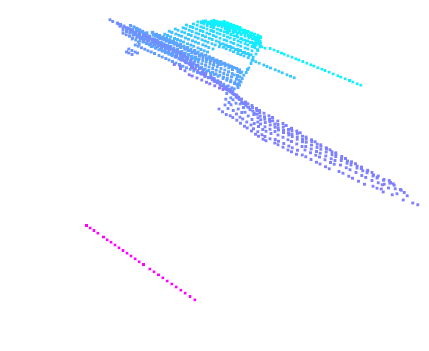}
		\centerline{Input}
	\end{minipage}
	\begin{minipage}[c]{0.23\linewidth}
		\includegraphics[width=\linewidth]{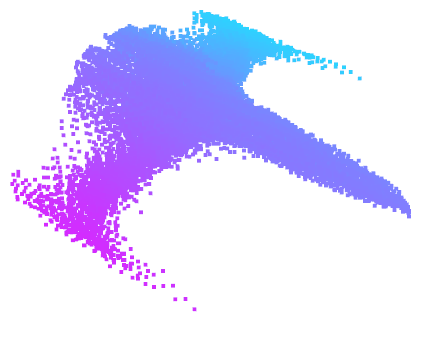}
		\centerline{PCN}
	\end{minipage}
	\begin{minipage}[c]{0.23\linewidth}
		\includegraphics[width=\linewidth]{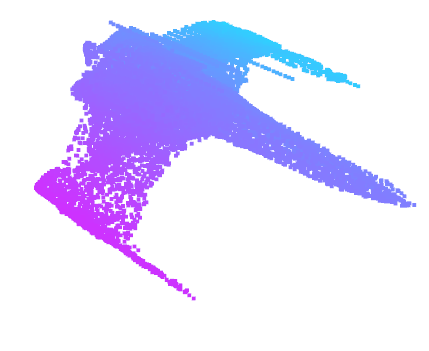}
		\centerline{MSN}
	\end{minipage}
	\begin{minipage}[c]{0.23\linewidth}
		\includegraphics[width=\linewidth]{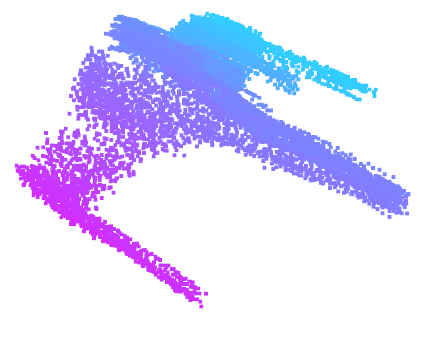}
		\centerline{GRNet}
	\end{minipage}
	\newline
	\begin{minipage}[c]{0.23\linewidth}
		\includegraphics[width=\linewidth]{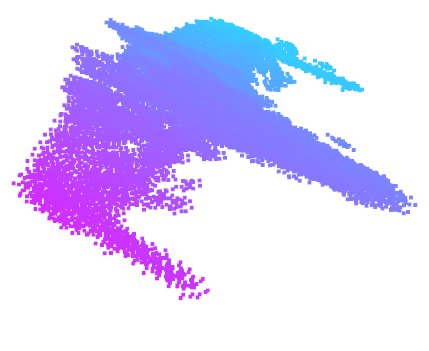}
		\centerline{CRN}
	\end{minipage}
	\begin{minipage}[c]{0.23\linewidth}
		\includegraphics[width=\linewidth]{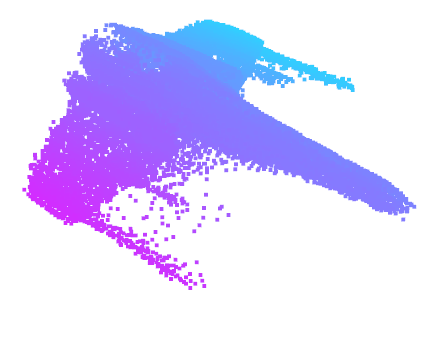}
		\centerline{NSFA}
	\end{minipage}
	\begin{minipage}[c]{0.23\linewidth}
		\includegraphics[width=\linewidth]{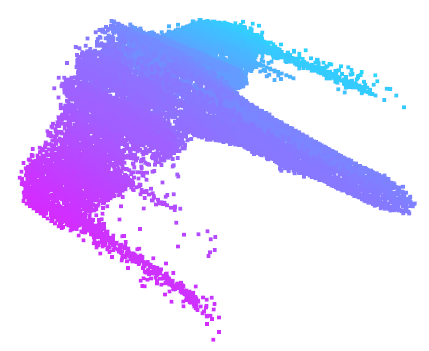}
		\centerline{Ours}
	\end{minipage}
	\begin{minipage}[c]{0.23\linewidth}
		\includegraphics[width=\linewidth]{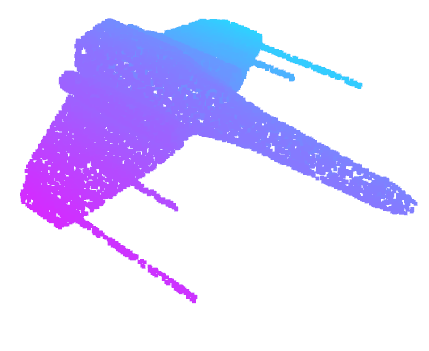}
		\centerline{GT}
	\end{minipage}
	\vspace{2mm}
	\caption{Completion results on ShapeNet. Each column from left to right is: input partial point cloud, results of PCN~\cite{yuan2018pcn}, MSN~\cite{liu2020morphing}, GRNet~\cite{xie2020grnet}, CRN~\cite{Wang_2020_CVPR}, NSFA~\cite{zhang2020detail}, ours, and the ground truth (GT).}
	\label{fig:failures}
\end{figure}

\subsection{Limitations and future work}

Although our results are better than others in general, occasionally we do observe that some results are not perfect. Our method outperforms others in recovering thin structures, but the completed thin structures may be noisy. For example, in Figure~\ref{fig:failures}, we observe that the guns mounted in the wings of the airplane are challenging to all methods to complete and recover. Other methods either miss the guns in the completion results (e.g., PCN, MSN, and GRNet) or produce highly noisy completion (e.g., CRN, NSFA). Our method is able to recover the four guns mounted in the wings but with some noisy points, which is not perfect. We will work on improving our framework to better deal with the recovery of small structures in the future.

\begin{table}[t]
  \centering
  \resizebox{0.8\linewidth}{!}{
    \begin{tabular}{l|c|c|c}
    \hline
    Method & Fidelity & MMD & Consistency \\
    \hline \hline
    PCN & 0.0278 & 0.0145 & 0.0137 \\
    CRN & 0.0276 & 0.0157 & 0.0186 \\
    GRNet & \underline{0.0149} & \textbf{0.0125} & \textbf{0.0089} \\
    NSFA & 0.0261 & 0.0154 & 0.0299 \\ \hline
    Ours & \textbf{0.0087} & \underline{0.0144} & \underline{0.0135}\\
    \hline
    \end{tabular}%
    }
    \vspace{1mm}
  \caption{Quantitative comparison results on LIDAR scanned cars from KITTI dataset. Best in \textbf{bold}, second best \underline{underlined}.}
  \label{tab:kitti}%
\end{table}%

\section{Conclusion}
\label{sec:conclusion}

To tackle the challenging dense 3D point cloud completion problem, we propose a novel framework that performs low-resolution recovery first, follows by a patch-wise noise-aware upsampling. Instead of decoding or regressing a complete and dense point cloud directly, which tends to lose geometric details and increase noise, our method achieves a high-fidelity dense point cloud completion through solving several easier subproblems including decoding a complete but sparse shape, iterative refinement, preserving trustworthy information by symmetrization, and patch-wise upsampling. The effectiveness of each component of our method is validated in the ablation studies. From the comparison evaluations, it is observed that our method outperforms the latest methods and establishes a new state of the art in point cloud completion. 

{\small
	\bibliographystyle{cvm}
	\bibliography{cvmbib}
}

\end{document}